\documentclass[10pt, a4paper]{article}
\usepackage[final]{lrec-coling2024} 

\usepackage{multirow}
\usepackage{multicol}
\usepackage{subcaption}
\usepackage{enumitem,kantlipsum}
\usepackage[normalem]{ulem}
\usepackage{tabularx}
\usepackage{color, colortbl}
\usepackage{todonotes} 
\usepackage{graphicx}
\usepackage{tabularx}
\usepackage{soul}
 
\usepackage{pbox}
\usepackage{booktabs} 
\usepackage{amssymb}
\usepackage{pifont}

\usepackage{tabu}
\usepackage{fontawesome}

\usepackage{makecell}
\usepackage{xcolor}
\usepackage{hyperref}
 \definecolor{darkblue}{rgb}{0, 0, 0.5}
  \hypersetup{colorlinks=true, citecolor=darkblue, linkcolor=darkblue, urlcolor=darkblue}

\usepackage{xstring}

\usepackage{color}

\usepackage{listings}
\usepackage{color}

\usepackage{pifont}
\usepackage{comment}

\usepackage{listings} 

\title{Zero- and Few-Shot Prompting with LLMs: A Comparative Study with Fine-tuned Models for Bangla Sentiment Analysis}

\name{Md. Arid Hasan$^1$, Shudipta Das$^2$, Afiyat Anjum$^2$, Firoj Alam$^3$,\\ \textbf{\large Anika Anjum$^2$, Avijit Sarker$^2$, Sheak Rashed Haider Noori$^2$}
}

\address{
$^1$SE+AI Research Lab, University of New Brunswick, Fredericton, Canada \\
$^2$Daffodil International University, Dhaka, Bangladesh\\
$^3$Qatar Computing Research Institute, Doha, Qatar\\ 
  \texttt{arid.hasan@unb.ca, fialam@hbku.edu.qa} \\
  }

\abstract{
The rapid expansion of the digital world has propelled sentiment analysis into a critical tool across diverse sectors such as marketing, politics, customer service, and healthcare. While there have been significant advancements in sentiment analysis for widely spoken languages, low-resource languages, such as Bangla, remain largely under-researched due to resource constraints. Furthermore, the recent unprecedented performance of Large Language Models (LLMs) in various applications highlights the need to evaluate them in the context of low-resource languages. In this study, we present a sizeable manually annotated dataset encompassing 33,606 Bangla news tweets and Facebook comments. We also investigate zero- and few-shot in-context learning with several language models, including Flan-T5, GPT-4, and Bloomz, offering a comparative analysis against fine-tuned models. Our findings suggest that monolingual transformer-based models consistently outperform other models, even in zero and few-shot scenarios. To foster continued exploration, we intend to make this dataset and our research tools publicly available\footnote{\url{https://github.com/AridHasan/MUBASE}}  to the broader research community. 
\\ \newline \Keywords{Sentiment, LLMs, Zero-shot, Few-shot, Bangla NLP, Low-resource language} 
}

\begin{document}

\maketitleabstract

\section{Introduction}
\label{sec:introduction}

Sentiment analysis is an influential sub-area of NLP that deals with sentiment, emotions, affect and stylistic analysis in language. There has been significant research effort for sentiment analysis due to its need in various fields, such as business, finance, politics, education, and services \cite{cui2023survey}. The analysis typically has been done on different types of content -- domains (news, blog posts, customer reviews, social media posts), modalities (textual and multimodal) \cite{hussein2018survey,dashtipour2016multilingual}. 
The surge in user-generated content on social media platforms has become a significant phenomenon, as individuals increasingly voice their opinions on a wide array of topics through comments and tweets. As a result, these platforms have garnered considerable research attention as valuable sources of data for sentiment analysis \cite{yue2019survey}. Leveraging such data resources \cite{dashtipour2016multilingual}, substantial progress has been achieved for the sentiment analysis in English. 
The advancements range from quantifying sentiment polarity to tackling more complex challenges like identifying aspects \cite{chen2022discrete}, multimodal sentiment detection \cite{liang2022msctd}, explainability \cite{cambria2022senticnet}, and multilingual sentiment analysis \cite{barbieri-etal-2022-xlm,galeshchuk-etal-2019-sentiment}.

\begin{figure}[h]
     \centering
     \includegraphics[width=0.48\textwidth ]{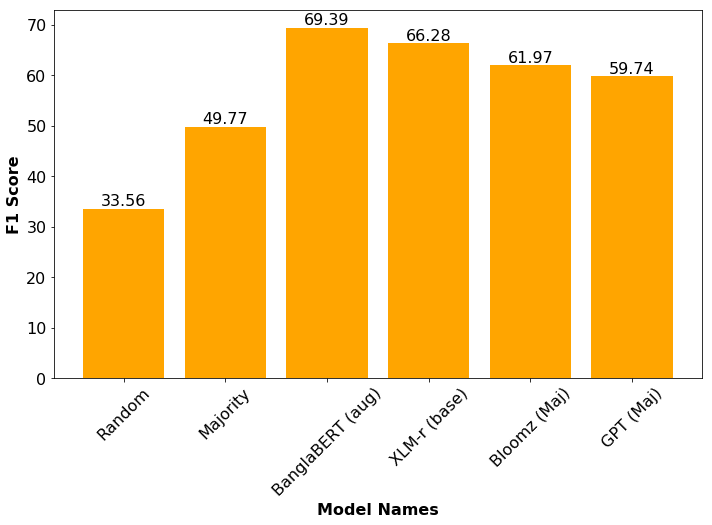}
     \caption{Performance comparisons with baselines (random and majority), fine-tuned models and LLMs (GPT and Bloomz).}
     \label{fig:model_comparison}
     \vspace{-0.4cm}
 \end{figure}
 
There has been a growing research interest over time in sentiment analysis for low-resource languages \cite{batanovic2016reliable,nabil2015astd,muhammad2023afrisenti}. Similar to other low-resource languages, the research for the sentiment analysis for Bangla has been limited \cite{islam2021sentnob,islam2023sentigold}.  
A study conducted by \citet{alam2021review} emphasized the primary challenges associated with Bangla sentiment analysis, specifically issues of duplicate instances in the data, inadequate reporting of annotation agreement, and generalization. These challenges were also highlighted in \cite{islam2021sentnob}, further emphasizing the need to address them for effective sentiment analysis in Bangla. To further facilitate sentiment analysis research in Bangla, we have created a multi-platform sentiment analysis dataset in this study. The dataset has undergone multiple rounds of pre-processing and validation to ensure its suitability for both sentiment analysis tasks and qualitative investigations.

We provide a comparative analysis using various pre-trained language models, including zero and few-shot settings with different fine-tuned models and LLMs as presented in Figure \ref{fig:model_comparison}. The analysis definitively demonstrates that LLMs surpass both random and majority baselines in performance, yet they fall short when compared to fine-tuned models.
More details are discussed in the \textit{Results} section.

Our contributions can be summarized as follows:
\begin{itemize} 
    \item Development of one of the largest manually annotated datasets for sentiment analysis.
    \item Investigation into zero-shot and few-shot learning using various LLMs. We are the first to provide such a comprehensive evaluation for Bangla sentiment analysis. 
    \item Comparative analysis of performance differences between in-context learning and fine-tuned models.
    \item Investigation of how different prompting variations affect performance in in-context learning. 
\end{itemize}

Based on our extensive experiments our findings are summarized below: 
\begin{itemize}
    \item Fine-tuned models yield better results compared to both zero- and few-shot in-context learning setups. 
    \item Fine-tuned models using monolingual text (BanglaBERT) demonstrate superior performance. 
    \item There is little to no performance difference between zero- and few-shot learning with the GPT-4 model.
    \item For the majority of zero- and few-shot experiments, BLOOMZ yielded better performance than GPT-4. 
    \item While BLOOMZ failed to predict the neutral class, GPT-4 struggled with positive class prediction.
\end{itemize}

The remainder of the paper is structured as follows: Section \textit{Related Work} provides an overview of relevant literature. The \textit{Dataset} section provides the details of the dataset used, along with an analysis of its contents. In Section \textit{Methodology}, we discuss the models and experiments. The \textit{Results and Discussion} section presents and discusses our findings. Lastly, Section \textit{Conclusion} provides concluding remarks.

\section{Related Work}
\label{sec:related_work}

In the realm of sentiment classification for Bangla, the current state-of-the-art research focuses on two key aspects: resource development and tackling model development challenges. Earlier work in this area has encompassed rule-based methodologies as well as classical machine learning approaches and recently the use of pre-trained models has received a wider attention.

\subsection{Datasets}
Over time there have been several resources developed including manual and semi-supervised labeling approaches \cite{6850712,alam2021review,islam2021sentnob,islam2023sentigold,kabir2023banglabook}. 
\citet{6850712} developed a dataset using semi-supervised approaches and trained models SVM and Maximum Entropy. The study of 
\citet{kabir2023banglabook} proposed an annotated sentiment corpus comprising 158,065 reviews collected from online bookstores. The annotations were primarily based on the rating of the reviews, with the majority (89.6\%) being in the positive class. The study also evaluated classical and BERT-based models for training and performance assessment. The skewness of this dataset makes it particularly challenging.
SentiGold~\cite{islam2023sentigold}\footnote{Note that this dataset is not publicly available.} is a well-balanced sentiment dataset containing 70K entries from 30 different domains. It was collected from various sources, including YouTube, Facebook, newspapers, blogs, etc., and labeled into five classes. The reported inter-annotator agreement is 0.88.

\citet{Rahman2018ABSA} labeled 5,700 instances as positive, negative, or neutral for two aspect-based sentiment analysis (ABSA) tasks, specifically extracting aspect categories and polarity. The authors curated two new datasets from the cricket and restaurant domains. 
\citet{islam2021sentnob} developed the SentNoB dataset, comprising 15,000 manually annotated comments collected from the comments section of Bangla news articles and videos across 13 diverse domains. The experimental findings using this dataset indicate that lexical feature combinations outperform neural models.


\subsection{Models}
Various classical algorithms have been employed in different studies for sentiment classification in Bangla. These include Bernoulli Naive Bayes (BNB), Decision Tree, Support Vector Machine (SVM), Maximum Entropy (ME), and Multinomial Naive Bayes (MNB) \cite{9084046,banik2018evaluation,chowdhury2019analyzing}.
\citet{islam2016supervised} developed a sentiment classification system for textual movie reviews in Bangla. The authors utilized two machine learning algorithms, Naive Bayes (NB) and SVM, and provided comparative results. Additionally, \citet{islam2016supervised} employed NB with rules for sentiment detection in Bengali Facebook statuses.

Deep learning algorithms have been extensively explored in the context of Bangla sentiment analysis~\cite{hassan2016sentiment,sharfuddin2018deep,tripto2018detecting,ashik2019data,DBLP:journals/corr/abs-2004-07807,sazzed2021improving,sharmin2021attention}.
In the study conducted by \citet{tripto2018detecting}, the authors utilized Long Short-Term Memory (LSTM) networks and Convolutional Neural Networks (CNNs) with an embedding layer to identify both sentiment and emotion in YouTube comments.
\citet{ashik2019data} conducted a comparative analysis of classical algorithms, such as Support Vector Machines (SVM), alongside deep learning algorithms, including LSTM and CNN, for sentiment classification of Bangla news comments.
\citet{DBLP:journals/corr/abs-2004-07807} integrated word embeddings into a Multichannel Convolutional-LSTM (MConv-LSTM) network, enabling the prediction of various types of hate speech, document classification, and sentiment analysis in the Bangla language.
Another aspect explored in sentiment analysis is the utilization of LSTM models due to the prevalence of romanized Bangla texts in social media. \citet{hassan2016sentiment,sharfuddin2018deep} employed LSTM models to design and evaluate their sentiment analysis models, taking into account the unique characteristics of romanized Bangla texts.

In the study conducted by \citet{hasan2020sentiment}, a comprehensive comparison was performed on various annotated sentiment datasets consisting of Bangla content from social media sources. The research investigated the effectiveness of both classical algorithms, such as SVM, and deep learning algorithms, including CNN and transformer models. Notably, the deep learning algorithm XLMRoBERTa exhibited superior performance with an accuracy of 0.671, surpassing the classical algorithm SVM, which achieved an accuracy of 0.581.

In a review article by \citet{alam2021review}, the authors investigated nine NLP tasks, including sentiment analysis. They reported that transformer-based models, particularly XLM-RoBERTa-large, are more suitable for Bangla text categorization problems than other machine learning techniques such as LSTM, BERT, and CNN.

\paragraph{Our Study:} We developed the largest MUltiplatform BAngla SEntiment (MUBASE) social media dataset, consisting of Facebook posts and tweets. Following the recommendations outlined in \citet{alam2021review}, we ensured that the dataset is clean, free of duplicates, and possesses high annotation quality with an annotation agreement score of $\kappa$=0.84. We have made the dataset publicly available\footnote{\url{https://github.com/AridHasan/MUBASE}} to the community. We conducted experiments that go beyond traditional approaches and smaller transformer-based models. Specifically, we investigated the effectiveness of advanced models such as Flan-T5, GPT-4, and BLOOMZ in both zero- and few-shot settings.

\section{Dataset}
\label{sec:dataset}

\subsection{Data Collection}
We collected tweets and comments from both Facebook posts and Twitter. To collect tweets, we focused on user accounts associated with the following news media sources: BBC Bangla, Prothom Alo, and BD24Live. For the comments from the Facebook posts, we selected public pages belonging to several news media outlets. Our selection of news media was based on the availability of a substantial number of comments. In total, we collected approximately 35,000 posts/comments associated with various Bangla news portals. Then we removed all the posts, which contains only emojis and URLs as well as duplicate data and filtered tweets while collecting through API. We also removed all the Banglish (Bangla text written in English alphabets) comments from our initial dataset. These filtering and duplicate-removal steps resulted in $33,606$ entries. In the rest of the paper, we will use the term post to refer to posts and comments. 

\begin{figure}[h]
     \centering
     \includegraphics[width=0.48\textwidth ]{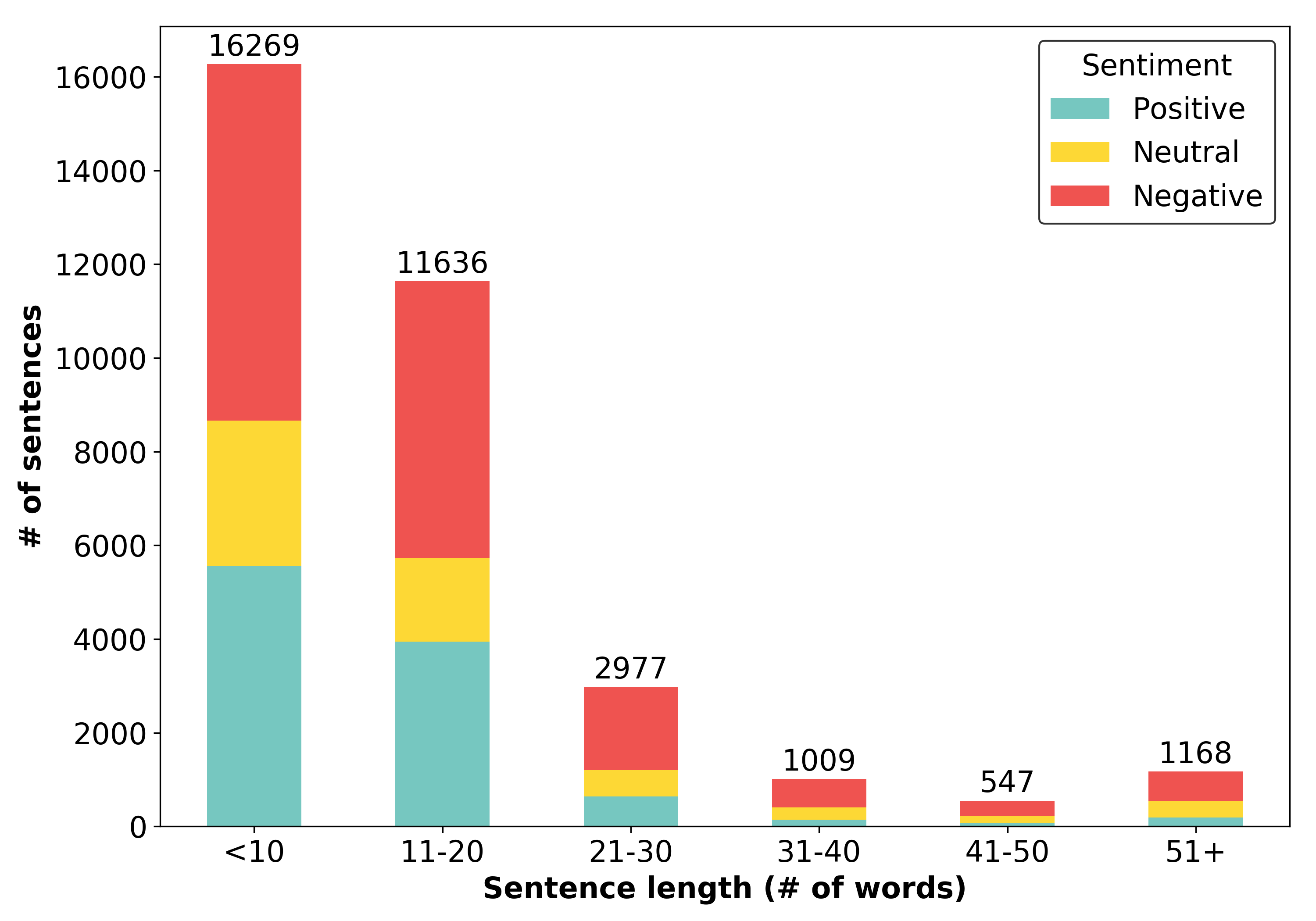}
     \caption{The distribution sentence length (number of words) associated with each sentiment label.}
     \label{fig:data_distribution}
 \end{figure}

Table \ref{tab:data_source_dist} presents the distribution of the number of posts and comments associated with each social media source. Our preliminary study reveals that Twitter users post both positive and negative sentiments, while showing fewer neutral expressions. On the other hand, Facebook users post more negative sentiments. Overall, the distribution of posts with negative sentiment is higher in the dataset.
We further analyzed the distribution of sentences by the number of words associated with each class label, as shown in Figure \ref{fig:data_distribution}. We created different ranges of sentence length buckets in order to understand and define the sequence length while training the transformer based models. It appears that more than 80\% of the posts lie within twenty words, which is expected with social media posts, as observed in previous studies \cite{alam2021humaid}.

\begin{table}[]
\centering
\begin{tabular}{lrrr}
\toprule
\multicolumn{1}{c}{\textbf{Class}} & \multicolumn{1}{c}{\textbf{Facebook}} & \multicolumn{1}{c}{\textbf{Twitter}} & \multicolumn{1}{c}{\textbf{Total}} \\ \midrule
Positive & 2,245 & 8,315 & 10,560 \\
Neutral & 4,866 & 1,331 & 6,197 \\
Negative & 9,078 & 7,771 & 16,849 \\
\textbf{Total} & 16,189 & 17,417 & 33,606 \\ \bottomrule
\end{tabular}
\caption{Class label distribution across different sources of the dataset.}
\label{tab:data_source_dist}
\end{table}

\subsection{Annotation}
To perform the annotation, we developed an annotation guideline based on previous studies \cite{mukta2021comprehensive}. For the Bangla sentiment polarity annotation, \citet{mukta2021comprehensive} proposed a classification with five labels: strongly negative, weakly negative, neutral, strongly positive, and weakly positive. However, due to the difficulty in distinguishing between strong and weak labels, we opted for a simplified approach with three labels: negative, neutral, and positive.

Each post was independently annotated by three annotators, all of whom are native speakers of Bangla. The annotators consisted of both male and female undergraduate students studying computer science. The final label for each post was determined based on the majority agreement among the annotators. However, in cases where there was disagreement among the annotators, a consensus meeting was organized to resolve any discrepancies and reach a final decision. Note that annotators are also the authors of the paper, hence, there has not been any payment for the annotation. 

\paragraph{Inter-Annotation Agreement}
The quality of the annotations was assessed by calculating the inter-annotator agreement. As mentioned previously, three annotators independently annotated each post, adhering to the provided annotation instructions. We calculated the Fleiss Kappa ($\kappa$) score and obtained a value of $0.84$, indicating a perfect agreement among the annotators.\footnote{Note that values of Kappa of and 0.81--1.0 correspond perfect agreement~\cite{landis1977measurement}.}


\subsection{Data Split}
For our experiments, we divided the dataset into training, development, and test sets, comprising 70\%, 10\%, and 20\% of the data, respectively. To ensure a balanced class label distribution across the sets, we employed stratified sampling \cite{sechidis2011stratification}. The distribution of the data split is provided in Table \ref{tab:data_dist}. 
The distribution in the table indicates a skew towards negative instances, followed by positive and neutral instances, suggesting that any analysis or model training based on this dataset may need to consider this imbalance.

\begin{table}[]
\centering
\begin{tabular}{lrrrr}
\toprule
\multicolumn{1}{c}{\textbf{Class}} & \multicolumn{1}{c}{\textbf{Train}} & \multicolumn{1}{c}{\textbf{Dev}} & \multicolumn{1}{c}{\textbf{Test}} & \multicolumn{1}{c}{\textbf{Total}} \\ \midrule
Positive & 7,342 & 1,126 & 2,092 & 10,560 \\
Neutral & 4,319 & 601 & 1,277 & 6,197 \\
Negative & 11,811 & 1,700 & 3,338 & 16,849 \\
\textbf{Total} & 23,472 & 3,427 & 6,707 & 33,606 \\ \bottomrule
\end{tabular}
\caption{Class label distribution of the dataset.}
\label{tab:data_dist}
\end{table}

\section{Methodology}
\label{sec:methodology}

\subsection{Data Pre-processing}
The content shared on social media is mostly noisy and includes emoticons, usernames, hashtags, URLs, invisible characters, and symbols. To clean the data, we removed the noisy portion (emoticons, usernames, hashtags, URLs, invisible characters, etc.) of the data. Then we applied tokenization and removed the stopwords from the data. Identifying usernames in Facebook posts is more challenging than in tweets. While tweets precede usernames with an `@' symbol, Facebook posts have no such distinguishing pattern. To address this, we removed English text from Facebook posts since most usernames are in English. However, for usernames in Bangla text, removal was challenging due to the absence of a consistent pattern or a comprehensive Bangla name dictionary.

\subsection{Evaluation Measures}
For the performance measure for all different experimental settings, we compute accuracy, and weighted precision, recall and F$_1$ score. We choose to use the weighted version of the metric as it takes into account class imbalance. 

\subsection{Training and Evaluation Setup}
For all experiments, except for LLMs (as detailed below), we trained the models using the training set, fine-tuned the parameters with the development set, and assessed the model's performance on the test set. For the LLMs, we accessed them through APIs. 

\subsection{Models}

We conducted our experiments using classical models as well as both small and large language models. It is worth noting that we follow the definitions of `small' and `large' models discussed in \cite{zhao2023survey}. The term `LLMs' refers to models encompassing tens or hundreds of billions of parameters.

\subsubsection{Baseline}
As baselines, we used both a majority (i.e., the class with the highest frequency) and a random approach. These methods have been widely used as baseline techniques in numerous studies, for example, \cite{rosenthal2019semeval}. 

\subsubsection{Classical Models}
While classical models such as SVM~\cite{platt1998} and Random Forest~\cite{breiman2001random} have been widely used in prior studies and remain in use in many low-resource production settings, we also wanted to assess their performance. To prepare the data for these models, we transformed the text into a tf-idf representation. During our experiments with SVM and RF, we used standard parameter settings: \textit{1)} used n-gram (1 to 5) and transformed them into TF-IDF, \textit{2)} for SVM we used the value of C =1, \textit{3)} and for the Random Forest we used number of trees as 200.

\subsubsection{Small Language Models (SLMs)}
Large-scale pre-trained transformer models (PLMs) have achieved state-of-the-art performance across numerous NLP tasks. In our study, we fine-tuned several of these models. These included the monolingual transformer model BanglaBERT \cite{bhattacharjee-etal-2022-banglabert} and multilingual transformers such as multilingual BERT (mBERT)~\cite{devlin2018bert}, XLM-RoBERTa (XLM-r)~\citep{conneau2019unsupervised}, BLOOMZ (560m and 1.7B parameters models)~\cite{muennighoff2022crosslingual}. We used the Transformer toolkit~\cite{Wolf2019HuggingFacesTS} for the experiment. Following the guidelines outlined in \citep{devlin2018bert}, we fine-tuned each model using the default settings over three epochs. Due to instability, we performed ten reruns for each experiment using different random seeds, and we picked the model that performed best on the development set. We provided the details of the parameters settings in Appendix \ref{sec:details_of_exp}. 

\subsubsection{GPT Embedding} 
For many downstream NLP tasks, embedding extracted from pre-trained models followed by fine-tuning a feed-forward network provided reasonable results and also a reasonable setup for a low-resource production setting. Hence, we wanted to see the performance of this setting. We first extract the embeddings using OpenAI's text-embedding-ada-002 model for each data split. We then fine-tune a feed-forward network on the embeddings extracted from the training set to train our model. Our feed-forward model utilizes the Rectified Linear Unit (ReLU) activation function. We have set the learning rate to 0.001 and the hidden layer size to 500. We validate our model using the validation set and finally, we evaluate the model using the test set.

\subsubsection{Large Language Models (LLMs)}
For the LLMs, we investigate their performance with in-context zero- and few-shot learning settings without any specific training. It involves prompting and post-processing of output to extract the expected content. Therefore, for each task, we experimented with a number of prompts, guided by the same instruction and format as recommended in the OpenAI Chat playground, and PromptSource~\cite{bach2022promptsource}. We used the following models: Flan-T5 (large and XL) \cite{chung2022scaling}, BLOOMZ (1.7B, 3B, 7.1B, 176B-8bit) \cite{muennighoff2022crosslingual} and GPT-4 \cite{openai2023gpt}. We set the temperatures to zero for all these models to ensure deterministic predictions. We used LLMeBench framework \cite{dalvi-etal-2024-llmebench} for the experiments, which provides seamless access to the API end-points and followed prompting approach reported in \cite{abdelali-etal-2024-larabench}.


\subsection{Prompting Strategy}
LLMs produce varied responses depending on the prompt design, which is a complex and iterative process that presents challenges due to the unknown representation of information within different LLMs. The instructions expressed in our prompts include both native (Bangla) and English languages with the input content in Bangla. 

\subsubsection{Zero-shot}
We employ zero-shot prompting, providing natural language instructions that describe the task and specify the expected output. This approach enables the LLMs to construct a context that refines the inference space, yielding a more accurate output. In Listing \ref{lst:zero_shot_gpt4}, we provide an example of a zero-shot prompt, emphasizing the instructions and placeholders for both input and label. Given that GPT-4 has the capability to play a role, therefore, we also provide a role for it as an ``expert annotator''
Along with the instruction we provide the labels to guide the LLMs and provide information on how the LLMs should present their output, aiming to eliminate the need for post-processing.

In our initial set of experiments with BLOOMZ, we observed that it did not respond as effectively to the same instructions as GPT-4. Therefore, we used more straightforward instructions for BLOOMZ, as illustrated in Listing \ref{lst:zero_shot_Bloomz}. For the other versions of BLOOMZ and Flan-T5, we used the same prompt as BLOOMZ. 


\begin{lstlisting}[caption={Zero-shot prompt example for GPT-4.}, label=lst:zero_shot_gpt4]
Instructions:
We would like you to analyze the sentiment of the following text. Based on the content of the text, please classify it as either "Positive", "Negative", or "Neutral". Provide only the label as your response.

text: {input_sample}

label:  

role: system, 
content: You are an expert annotator. Your task is to analyze the text and identify sentiment polarity.
\end{lstlisting}


\begin{lstlisting}[caption={Zero-shot prompt example for BLOOMZ.}, label=lst:zero_shot_Bloomz]
Instructions:
Label the following text as Neutral, Positive, or Negative. 
Provide only the label as your response.

text: {input_sample}

label:
\end{lstlisting}

\subsubsection{Few-shot}
The seminal work by \citet{brown2020language} demonstrated that few-shot learning offers superior performance when compared to the zero-shot learning setup. This has also been proven by numerous benchmarking studies (e.g., \cite{ahuja2023mega}). In our study, we conducted few-shot experiments using GPT-4 and BLOOMZ. For few-shot learning, we selected examples from the available training data. 
We used maximal marginal relevance-based (MMR) selection to construct example sets that are deemed relevant and diverse \cite{carbonell1998use}. This approach has been demonstrated as a successful method for selecting few-shot examples by \citet{ye2022complementary}. The MMR technique calculates the similarity between a test example and the training dataset, subsequently selecting $m$ examples (shots). This selection was performed on top embeddings obtained from multilingual sentence-transformers \cite{reimers2019sentence}. We chose to use 3- and 5-shots to optimize the API cost. The effect of $m$ setting will be explored in our future study. Note that our experiments of few-shot with BLOOMZ were worse than zero-shot, which might require further investigation. Therefore, in this study, we do not further discuss the BLOOMZ experiments with few-shot. 

In Listing \ref{lst:few_shot_gpt4}, we present an example of a few-shot prompt for GPT-4. The few-shot prompt distinguishes itself from the zero-shot in several ways:
\begin{itemize}
    \item We provided additional information for the role, 
    \item We simplified the instructions, and
    \item We included $m$-shot examples.
\end{itemize}

Our choices of prompts were based on our extensive experiments on similar tasks. 



\begin{lstlisting}[caption={Few-shot prompt example for GPT-4.}, label=lst:few_shot_gpt4]
Instructions:
Annotate the "text" into "one" of the following categories: "Positive", "Negative", or "Neutral".
Here are some examples:
Example 1: 
text: {input_example}
label: {input_label}

Example 2:
...

text: {input_sample}

label:  

role: system,
content: As an AI system, your role is to analyze text and classify them as 'Positive', 'Negative' or 'Neutral'. Provide only label and in English.
\end{lstlisting}

\section{Result and Discussion}
\label{sec:result_discussion}





\begin{table}[ht]
\centering
\setlength{\tabcolsep}{4pt}
\resizebox{0.45\textwidth}{!}{
\begin{tabular}{lrrrr}
\toprule
\multicolumn{1}{c}{\textbf{Exp}} & \multicolumn{1}{c}{\textbf{Acc}} & \multicolumn{1}{c}{\textbf{P}} & \multicolumn{1}{c}{\textbf{R}} & \multicolumn{1}{c}{\textbf{F1}} \\ \midrule
\multicolumn{5}{c}{\textbf{Baseline}} \\ \midrule
Random  & 33.56 & 38.31 & 33.56 & 33.56 \\
Majority & 49.77 & 24.77 & 49.77 & 49.77 \\ \midrule
\multicolumn{5}{c}{\textbf{Classic Models}} \\ \midrule
SVM & 55.81 & 53.33 & 55.81 & 52.39 \\
RF & 56.75 & 54.61 & 56.75 & 52.62 \\ \midrule
\multicolumn{5}{c}{\textbf{Fine-tuning}} \\ \midrule
Embedding (GPT) & 57.79 & 57.30 & 57.79 & 57.46 \\ 
Bloomz-560m & 61.71 & 63.08 & 61.97 & 63.08 \\
Bloomz-1.7B & 61.16 & 59.76 & 61.16 & 59.95 \\
BERT-m & 64.95 & 64.92 & 64.95 & 64.90 \\
XLM-r (base) & 66.63 & 66.24 & 66.63 & 66.28 \\
XLM-r (large) & 66.33 & 65.63 & 66.33 & 65.79 \\ 
BanglaBERT & 69.08 & 67.61 & 69.08 & 67.98 \\
BanglaBERT\textbf{*} & 70.33 & 69.13 & 70.33 & \textbf{69.39} \\
\midrule
\multicolumn{5}{c}{\textbf{Zero- and Few-shot on LLMs}} \\ \midrule
\multicolumn{5}{c}{\textbf{Open Models - 0-shot}} \\ \midrule
Flan-T5 (large) & 41.28 & 20.23 & 13.77 & 20.23 \\
Flan-T5 (xl) & 49.42 & 29.46 & 18.18 & 29.46 \\
Bloomz-1.7B  & 58.33 & 49.38 & 58.33 & 50.38 \\
Bloomz-3B   & 59.73 & 50.98 & 59.73 & 51.53 \\
Bloomz-7.1B  & 62.83 & 50.92 & 62.83 & 56.24 \\
Bloomz 176B (8bit) & 61.84 & 51.16 & 61.84 & 55.54 \\ \midrule
Bloomz Majority & 61.97 & 51.32 & 61.97 & 61.97 \\ \midrule
\multicolumn{5}{c}{\textbf{Closed Models - $m$-shot}} \\ \midrule
GPT-4: 0-Shot & 60.21 & 61.65 & 60.21 & 59.99 \\
GPT-4: 0-Shot (BN inst.) & 60.70 & 61.71 & 60.70 & 59.96 \\
GPT-4: 3-Shot & 60.40 & 63.88 & 60.40 & 60.74 \\
GPT-4: 5-Shot & 60.95 & 63.83 & 60.95 & 61.17 \\ \midrule
GPT-4 Majority & 59.74 & 63.26 & 59.74 & 59.74 \\
\bottomrule
\end{tabular}
}
\caption{Performance of different sets of experiments. \textbf{*} indicates trained on combined MUBASE, SentiNoB\cite{islam2021sentnob}, and \citet{alam2021review}. BN Ins. refers that instruction is provided in the native Bangla language.}
\label{tab:results}
\end{table}

In Table \ref{tab:results}, we reported the results of our experiments. 
\subsection{Comparison with Baselines:} All experimental setup outperformed random and majority baselines except Flan-T5. We calculate the random baseline by assigning a label to each test instance randomly, with the choice of label present in the training set. For the majority baseline, we identify the most common class within the training set and assign this class as the prediction for every instance in the test dataset, subsequently computing the performance.

\subsection{Performance of Classic Models} The performance of the SVM and Random Forest better than baseline, however, worse than others except Flan-T5. Comparatively, the SVM and Random Forest models exhibit similar performance levels.

\subsection{Fine-tuning}
Fine-tuned models consistently outperform across various settings. Results using GPT embeddings are superior to classical models, though not as effective as some other approaches. Although multilingual models such as BERT-m, XLM-r, and BLOOMZ show promising direction, however, models trained on monolingual text ultimately achieve superior performance.

Given the superior performance of monolingual models across various settings, we chose to augment our training data. By integrating the SentiNoB training set with the MUBASE training set and fine-tuning with BanglaBERT, we managed to boost performance by an additional 1.41\% of F1. 

When comparing the smaller BLOOMZ model (560m) to the larger one (1.7B), the smaller model performs better. This suggests that more training data might be required to effectively train such a large model. A similar pattern is observed with the XLM-r model when comparing its base and large versions.

\subsection{Zero- and Few-shot Prompt-Based Results}

\subsubsection{BLOOMZ:} As can be seen in Table \ref{tab:results}, the performance of zero- and few-shot approaches is promising, though there is a significant difference compared to the best monolingual fine-tuned transformer-based model. When comparing different parameter sizes of BLOOMZ, we observe that performance increases from 1.7B to 7.1B. However, we see a lower performance with BLOOMZ 176B compared to 7.1B, which might be due to the 8-bit precision. 

\subsubsection{Ensemble:} We hypothesized that predictions from different models might vary, and an ensemble of their outputs might provide better results. Therefore, we opted to use a majority-based ensemble method, resulting in a 5.73\% improvement in weighted F1. 

\subsubsection{GPT4:} The performance of GPT-4 is higher than that of other LLMs. Our experiments with different types of prompting did not yield a clear improvement, as can be seen in Table \ref{tab:results}. While prior studies on other tasks and languages showed a clear performance gain with a few-shot setup, in our study, we did not find such a gain, only slight differences in precision. Therefore, our future studies will include further investigation of few-shot learning setups. 

Our experiments revealed that native language instructions achieved performance comparable to that of English instructions. This indicates the potential for using native language prompts for Bangla sentiment analysys.

While the ensemble of different BLOOMZ settings improved performance, however, it did not help for GPT-4.

\subsubsection{Error Analysis on the Output of Prompts:}
Further analysis the results of the LLMs outputs we observed that {\em(i)} Flan-T5 (xl) labeled only five posts as \textit{negative}, and Flan-T5 (large) labeled only 45 posts as \textit{negative}, {\em(ii)} BLOOMZ completely failed to label posts as \textit{neutral}, and {\em(iii)} GPT-4 struggled to predict \textit{positive} class. 


In Table \ref{tab:test_performance_bert-bn}, we present the class-wise classification performance. The results indicate a higher F1 score for the negative class in comparison to the neutral and positive classes. This performance aligns with the class label distribution detailed in Table \ref{tab:data_dist}, where $\sim$50\% of the data corresponds to the negative class, followed by $\sim$31\% for the positive class and $\sim$18\% for the neutral class.

\begin{table}[ht]
\centering
\begin{tabular}{@{}lrrr@{}}
\toprule
{\textbf{Class}} & {\textbf{P}} & {\textbf{R}} & {\textbf{F1}} \\
\midrule
Negative       & 0.7512      & 0.7771   & 0.7640  \\
Neutral       & 0.4616      & 0.3156   & 0.3749  \\
Positive       & 0.6871      & 0.7820   & 0.7315  \\ \midrule
\bottomrule
\end{tabular}
\caption{Detail results on the test set with the model trained using BERT-bn.}
\label{tab:test_performance_bert-bn}

\end{table}
\section{Conclusion}
In this study, we present our evaluation of LLMs using zero and few-shot prompting. We offer a detailed comparison with fine-tuned models. Our experiments were conducted on a newly developed dataset named ``MUBASE'', for which we provide an in-depth analysis. Our results indicate that while LLMs represent a promising research direction, the smaller versions of fine-tuned pre-trained models outperform them. The performance of LLMs suggests that sentiment analysis in a new domain is feasible with reasonable accuracy without the need to develop a new dataset or train a new model. Future research directions include using other recently released datasets and providing a comparative analysis with LLMs. Additionally, further study on few-shot learning represents another promising research avenue.


\section*{Ethics Statement}
Our dataset may include posts with negative statements, so we advise model developers to use it cautiously. Using this dataset without careful consideration could lead to misleading results for the audience. Although the dataset's annotation is subjective, we have attempted to minimize this by obtaining annotations from multiple annotators.

%

%

\bibliographystyle{lrec-coling2024-natbib}
\bibliography{bib/main}



\appendix
\section{Details of the experiments}
\label{sec:details_of_exp}
For the experiments with transformer models, we adhered to the following hyper-parameters during the fine-tuning process. Additionally, we have released all our scripts for the reproducibility.

\begin{itemize}[nosep]
    \item Batch size: 8;
    \item Learning rate (Adam): 2e-5;
    \item Number of epochs: 10;
    \item Max seq length: 256.
\end{itemize}

\textbf{Models and Parameters:}
\begin{itemize}[nosep]
    \item \textbf{BanglaBERT} (csebuetnlp/banglabert):L=12, H=768, A=12, total parameters: 110M; where \textit{L} is the number of layers (i.e.,~Transformer blocks), \textit{H} is the hidden size, and \textit{A} is the number of self-attention heads; (110M);
    \item \textbf{XLM-RoBERTa} (xlm-roberta-base): L=24, H=1027, A=16; the total number of parameters is 355M.
    \item \textbf{BLOOMZ} (bigscience/bloom-560m): L=24, H=1024, A=16; the total number of parameters is 560M.
    \item \textbf{BLOOMZ} (bigscience/bloom-1b7): L=24, H=2048, A=16; the total number of parameters is 1.7B.




\end{itemize}



\end{document}